\documentclass[runningheads,a4paper]{llncs}

\usepackage{amssymb}
\usepackage{graphicx}
\usepackage{caption}
\usepackage{subfigure}
\usepackage{url}

\urldef{\mailsa}\path|{alfred.hofmann, ursula.barth, ingrid.haas, frank.holzwarth,|
\urldef{\mailsb}\path|anna.kramer, leonie.kunz, christine.reiss, nicole.sator,|
\urldef{\mailsc}\path|erika.siebert-cole, peter.strasser, lncs}@springer.com|
\newcommand{\keywords}[1]{\par\addvspace\baselineskip
\noindent\keywordname\enspace\ignorespaces#1}

\usepackage[breaklinks=true,hyperindex=true,colorlinks=true,bookmarks=false,citecolor=blue]{hyperref}
\begin{document}

\mainmatter  

\title{On the Exploration of Convolutional Fusion Networks for Visual Recognition}

\titlerunning{Convolutional Fusion Networks for Visual Recognition}

%
%
%
\author{Yu Liu, Yanming Guo, and Michael S. Lew}

\institute{LIACS Media Lab, Leiden University, Leiden, The Netherlands\\
{\tt\small \{y.liu, y.guo, m.s.lew\}@liacs.leidenuniv.nl}
}
%
%

\toctitle{Lecture Notes in Computer Science}
\tocauthor{Authors' Instructions}
\maketitle

\begin{abstract}
Despite recent advances in multi-scale deep representations, their limitations
are attributed to expensive parameters and weak fusion modules.
Hence, we propose an efficient approach to fuse multi-scale deep representations, called convolutional fusion networks (CFN).
Owing to using 1$\times$1 convolution and global average pooling,
CFN can efficiently generate the side branches while adding few parameters.
In addition, we present a locally-connected fusion module, which can learn adaptive weights
for the side branches and form a discriminatively fused feature.
CFN models trained on the CIFAR and ImageNet datasets demonstrate remarkable improvements over the plain CNNs.
Furthermore, we generalize CFN to three new tasks, including scene recognition,
fine-grained recognition and image retrieval.
Our experiments show that it can obtain consistent improvements towards the transferring tasks.
\end{abstract}

\section{Introduction}
Since their repeated success in ImageNet classification~\cite{Alexnet2012,VGGnet2015,googlenet2015,ResNet2015},
deep convolutional neural networks (CNNs) have contributed much to
computer vision and the wider research community around it.
CNN features can be used for many visual recognition
tasks, and obtain top-tier performance~\cite{generic2015}.
Moreover, some works~\cite{Multilayer2014,treasure2015,MPP2015} begin capturing
complementary features from intermediate layers.
However, their methods mainly make use of one off-the-shelf model trained on the ImageNet dataset~\cite{Imagenet2015}, but not to
train a new network that can integrate intermediate layers.
Instead, recent work~\cite{DAG2015} trains a multi-scale architecture for scene recognition
at the expense of increasing algorithm complexity.

Hence, we propose to train an efficient fusion architecture to integrate intermediate layers for visual recognition.
Our architecture is called convolutional fusion networks (CFN), which mainly consists of
three characteristics: (1) \emph{Efficient side outputs}: we add few parameters to generate new side branches
due to using efficient 1$\times$1 convolution and global average pooling~\cite{NIN2014}.
(2) \emph{Early fusion and late prediction}: in contrast to~\cite{DAG2015},
we present an ``early fusion and late prediction'' strategy.
It can not only reduce the number of parameters, but also produce a richer image representation.
(3) \emph{Locally-connected fusion}: in the fusion module, we propose making use of a locally-connected layer
to learn adaptive weights (importance) for the side outputs.
To the best of our knowledge, this is the first attempt to apply a locally-connected layer to a fusion module.

In a nutshell, our contributions can be summarized as follows. First, an efficient fusion
architecture is presented to provide promising insights into efficiently exploiting multi-scale deep features.
Second, we train CFN on the CIFAR and ImageNet 2012 datasets, and evaluate its efficiency and effectiveness.
Experimental results demonstrate the superiority of CFN over the plain CNN.
Third, we generalize the CFN model to other new tasks, including
scene recognition, fine-grained recognition and image retrieval.
Results show that CFN can consistently achieve significant improvements
on these transferring tasks.

\section{Related work}
In this section, we summarize existing approaches that focus on intermediate layers
in the following three aspects.

\textbf{Employment of intermediate layers.}
In CNNs, intermediate layers can capture complimentary information to the top-most layers.
For example, Ng, et al.~\cite{Ng2015} employed features from different intermediate layers and encoded them
with VLAD scheme. Similarly, Cimpoi, et al.~\cite{banks2015} and Wei, et al.~\cite{DSP2015} made use of
Fisher Vectors to encode intermediate activations.
Moreover, Liu, et al.~\cite{treasure2015} and Babenko, et al.~\cite{Babenko2015}
aggregated several intermediate activations
and generated a more discriminative and expensive image descriptor.
Based on intermediate layers, these methods are able to achieve promising performance on their tasks,
as compared to using the fully-connected layers.

\textbf{Intermediate supervision.}
Considering the importance of intermediate layers,
Lee, et al.~\cite{DSN2015} proposed the deeply supervised nets, which
imposed additional supervision to guide the intermediate layers earlier, rather than the standard approach
of only supervising the final prediction.
Similarly, GoogLeNet~\cite{googlenet2015} created two extra branches from the intermediate layers
and supervised them jointly.
However, these approaches do not explicitly fuse the outputs of intermediate layers.

\textbf{Multi-scale fusion (or skip connections).}
To incorporate intermediate outputs explicitly during training, multi-scale fusion is
presented to train multi-scale deep neural networks~\cite{DAG2015,FCN2015,HED2015}.
A similar work in~\cite{DAG2015} builded a DAG-CNNs model that summed up the multi-scale predictions from intermediate layers.
However, DAG-CNNs required processing a large number of additional parameters.
In addition, its fusion module (i.e. sum-pooling) failed to consider the importance of side branches.
However, our CFN can learn adaptive weights for fusing side branches, while adding few parameters.

\section{Proposed approach}
In this section, we introduce the architecture of CFN and its training procedure.

\clearpage

\subsection{Architecture}
\begin{figure}[t]
\setlength{\abovecaptionskip}{0.cm}
\captionsetup{belowskip=-10pt}
\centering
\includegraphics[width=12cm, height=3.5cm]{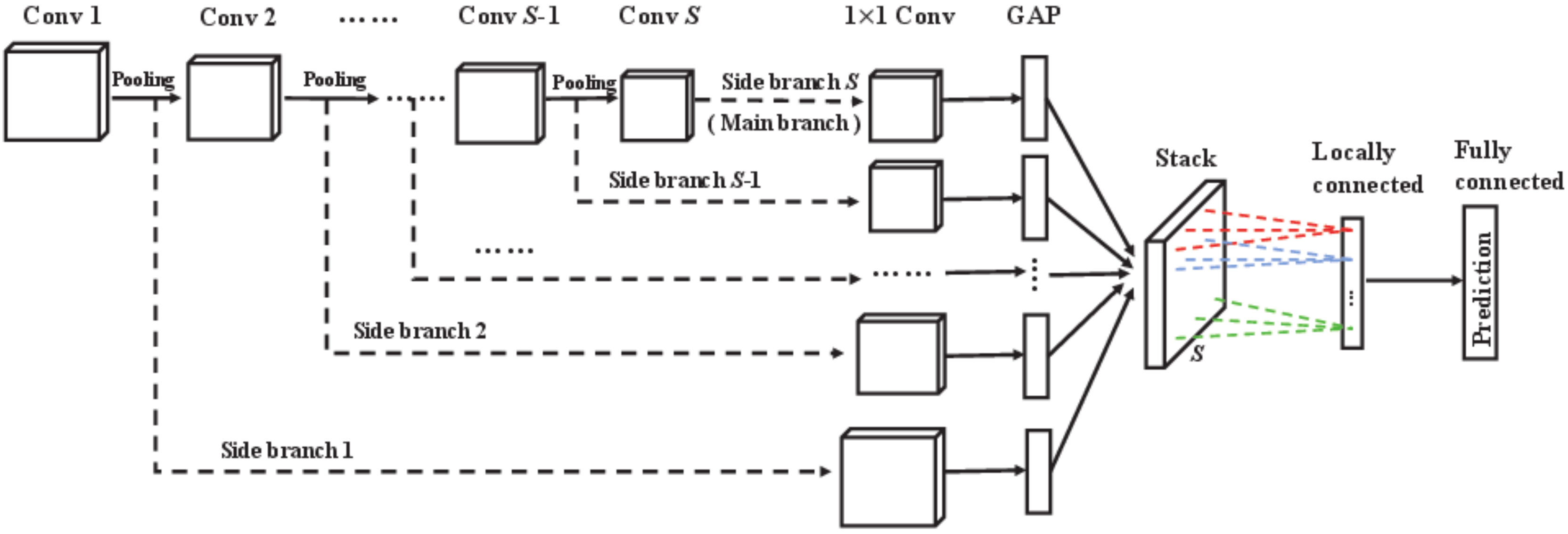}
\caption{The general pipeline of convolutional fusion networks (best viewed in zoom in).
         The side branches start from the pooling layers and consist of a 1$\times$1 convolution
         and global average pooling.
         All side outputs are then stacked together.
         A locally-connected layer is used to adaptively learn adaptive
         weights for the side outputs.
         Finally, the fusion feature is fed to the following fully-connected layer
         that is used to make a final prediction.}
\label{fig:DFN}
\end{figure}

Similar to ~\cite{NIN2014,googlenet2015,ResNet2015}.
we use 1$\times$1 convolutional layer and global average pooling at the top layers,
To reduce the number of parameters in a plain CNN model.
Based on a plain CNN, we develop our convolutional fusion networks, as illustrated in Fig.~\ref{fig:DFN}.
Overall, our CFN manly consists of three characteristics that will be described in the following.

\textbf{Efficient side outputs.}
Instead of using the fully-connected layers, CFN efficiently generates the side branches from the
intermediate layers while adding few parameters.
First, the side branches are grown from the pooling layers by inserting 1$\times$1 convolution layers like the main branch.
All 1$\times$1 convolutional layers must have the same number of channels
so that they can be integrated together.
Then, global average pooling is performed over the 1$\times$1 convolutional maps to
obtain one-dimensional feature vector, called GAP feature here.
Notably, we can also consider the full depth main branch as a side branch.

Assume that there are $S$ of side branches in total and the last side branch (i.e. $S$-th) indicates the main branch.
We notate $h_{i,j}^{(s)}$ as the input of 1$\times$1 convolution in the $s$-th side branch,
where $s=1,2,\dots,S$ and $(i,j)$ is the spatial location across feature maps.
As 1$\times$1 convolution has $K$ of channels,
its output associated with the $k$-th kernel, denoted as $f_{i,j,k}^{(s)}$,
where $k=1,\dots,K$.
Next, let $H^{(s)}$ and $W^{(s)}$ be the height and width of features maps derived from the $s$-th 1$\times$1 convolution.
Thereby, global average pooling performed over the feature map $f_{k}^{(s)}$ is calculated by
\begin{equation}
g_{k}^{(s)} = \frac{1}{H^{(s)}W^{(s)}}\sum_{i=1}^{H^{(s)}}\sum_{j=1}^{W^{(s)}} f_{i,j,k}^{(s)},
\end{equation}
Where $g_{k}^{(s)}$ is the $k$-th element in the $s$-th GAP feature vector.
Thus, we can notate $g^{(s)}=[g_{1}^{(s)},\dots,g_{K}^{(s)}]$, a $1 \times K$ dimensional vector,
as the whole GAP feature from the $s$-th side branch.
Recall that $g^{(S)}$ represents the GAP feature from the full depth main branch.
\begin{figure}[t]
\setlength{\abovecaptionskip}{0.cm}
\captionsetup{belowskip=-10pt}
\centering
\subfigure[] {\includegraphics[height=2cm]{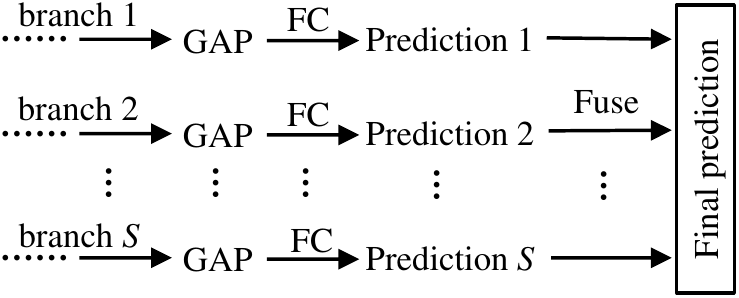} \label{fig:EPLF}} \hspace{0.8cm}
\subfigure[] {\includegraphics[height=2cm]{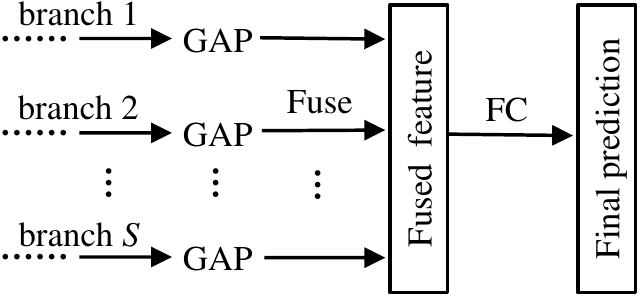} \label{fig:EFLP}}
\caption{Comparison between EPLF and EFLP. (a) The pipeline of EPLF strategy;
         (b) The pipeline of EFLP strategy.
        }
\label{fig:earlylate}
\end{figure}

\textbf{Early fusion and late prediction.}
Considering when to fuse the side branches,
related work~\cite{DAG2015,HED2015} used an ``early prediction and late fusion'' (EPLF) strategy.
In contrast to EPLF~\cite{DAG2015}, in which a couple of FC layers are added,
we present an opposite strategy called ``early fusion and late prediction'' (EFLP).
EFLP can fuse the GAP features from the side outputs and obtain a fused feature.
Then, one fully-connected layer following the fused feature is
used to estimate the final prediction.
Figure~\ref{fig:earlylate} shows the comparison between EPLF and EFLP.
As compared to EPLF, EFLP consumes less parameters due to using only one fully-connected layer.
We assume that each fully-connected layer has $C$ units that correspond to the number of object categories in the dataset.
The fusion module has $W_{fuse}$ of parameters.
Quantitatively, we can compare the parameters (i.e. weights and bias) between EFLP and EPLF by
\begin{equation}
W_{EFLP} =  S (C+1) + W_{fuse} < W_{EPLF} = S K (C+1) + W_{fuse}.
\end{equation}

More importantly, the fused feature in EFLP can be extracted as a richer image representation,
compared with the widely-used fc6 and fc7~\cite{Alexnet2012,VGGnet2015}.
The fused feature could be transferred from generic to specific vision recognition tasks.
However, EPLF fails to specify which feature can serve as a good representation.
Additionally, EFLP can achieve the same accuracy as EPLF, though EPLF consumes more parameters.

\textbf{Locally-connected fusion.}
Another significant component in CFN is its fusing the branches based on a locally-connected (LC) layer.
Owing to its no-sharing filters over spatial dimensions, LC layer can learn different weights
in each local field~\cite{Local2010}.
We aim to make use of a LC layer to learn adaptive weights (or importance) for the side branches,
and to generate the fused feature.
As we know, this is the first attempt to apply a locally-connected layer to a fusion module.
The detail computation are introduced as follows.
\begin{figure}[t]
\setlength{\abovecaptionskip}{0.cm}
\captionsetup{belowskip=-10pt}
\centering
\subfigure[No weights]{\includegraphics[width=3.5cm,height=2.5cm]{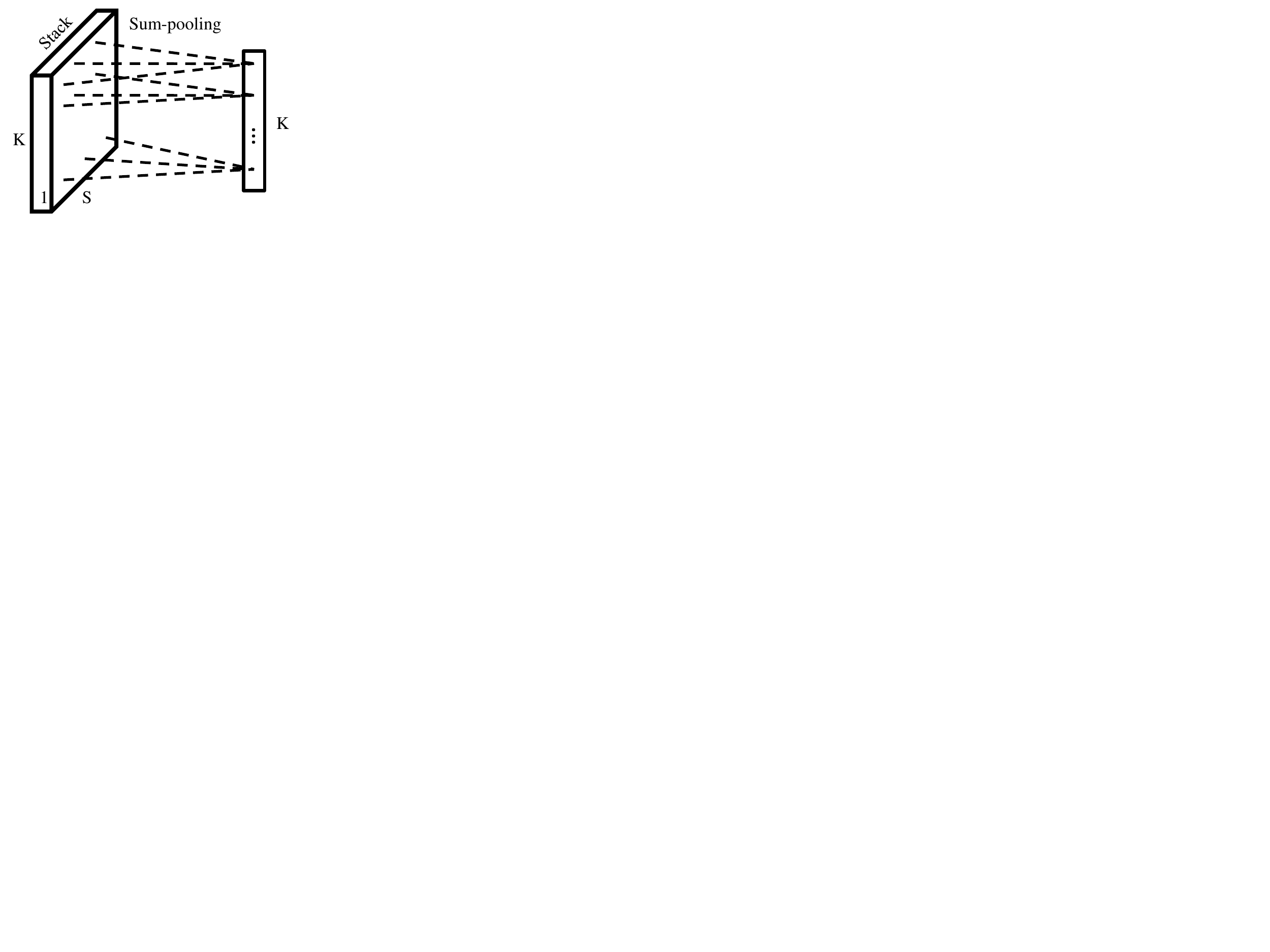} \label{fig:local_a}} \hspace{0.4cm}
\subfigure[Sharing weights]{\includegraphics[width=3.5cm,height=2.5cm]{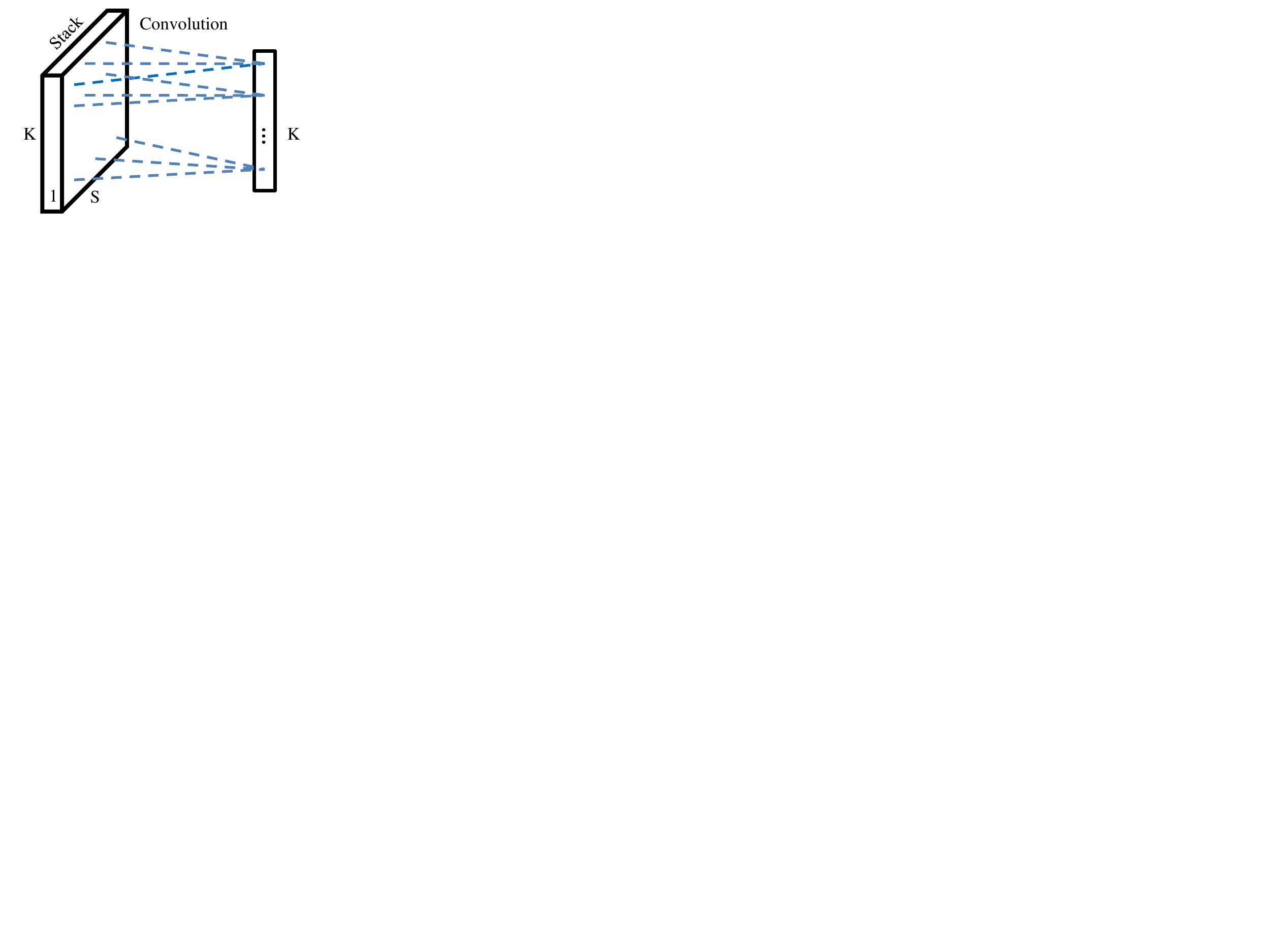} \label{fig:local_b}} \hspace{0.4cm}
\subfigure[No-sharing weights]{\includegraphics[width=3.5cm,height=2.5cm]{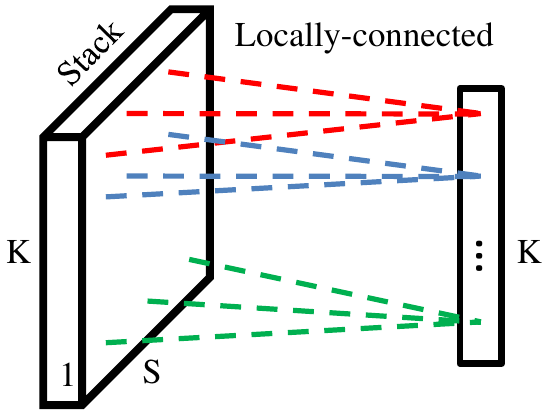} \label{fig:local_c}}
\caption{Comparison of three fusion modules (best viewed in color).
         Left: Sum-pooling fusion has no weights;
         Middle: Convolution fusion learns sharing weights over spatial positions, as drawn in the same color;
         Right: Locally-connected fusion learns no-sharing weights over spatial positions, as drawn in different colors.
             To learn element-wise weights, we use 1$\times$1 local field.
        }
\label{fig:local}
\end{figure}

At first, we stack GAP features together (from $g^{(1)}$ to $g^{(S)}$), and form a
stack layer $G$ with size of $1 \times K \times S$, see Fig.~\ref{fig:DFN}.
The $s$-th feature map of $G$ is $g^{(s)}$.
Then, one LC layer which has $K$ of no-sharing filters is convolved over $G$. Each filter has 1$\times$1$\times S$ kernel size.
As a result, LC can learn adaptive weights for different elements in the GAP features.
Here, the fused feature convolved by LC also has $1 \times K$ shape, denoted as $g^{(f)}$.
Each element in $g^{(f)}$ can be computed via
\begin{equation}
g_{i}^{(f)} = \sigma \left( \sum_{j=1}^{S} W^{(f)}_{i,j} \cdot g_{i}^{(j)} + b^{(f)}_{i} \right),
\end{equation}
where $i=1,2,\dots,K$; $\sigma$ indicates the activation function (i.e. ReLU).
$W^{(f)}_{i,j}$ and $b^{(f)}_{i}$ represent the weights and bias.
The number of parameters in the LC fusion is $K \times (S+1)$.
These additional parameters benefit adaptive fusion
while do not need any manual tuning.

To be clear, Figure~\ref{fig:local} compares LC fusion with other simple fusion methods.
In Fig.~\ref{fig:local_a}, the sum-pooling fusion simply sums up the side outputs without learning any weights.
In Fig.~\ref{fig:local_b}, the convolution fusion can learn only one sharing filter
over the whole spatial dimensions (as drawn in the same blue color).
On the contrary, LC can learn independent weights over each local field (i.e. 1$\times$1$\times$S size),
as drawn in different colors in Fig.~\ref{fig:local_c}.
Although LC fusion has a little more parameters than the sum-pooling and convolution fusion,
these parameters are nearly negligible as compared to the whole network parameters.

\subsection{Training}
Since CFN has efficient forward propagation and backward propagation,
it can maintain the ease of training as similar to CNN.
Assume that $W$ indicates the set of all parameters learned in the CFN (including the LC fusion weights),
and $\mathcal{L}$ is the total loss cost during training.
To minimize the total loss, the partial derivative of the loss with respect
to any weight will be recursively computed by the chain rule during the backward propagation~\cite{LeNet1990}.
Since the main components in our CFN model are the side branches, we will induce
the detail computations of their partial derivatives.
For notational simplicity, we consider each image independently in the following.

First, we compute the gradient of the loss cost with respect to the outputs of the side branches.
As an example of the $s$-th side branch,
the gradient of $\mathcal{L}$ with respect to
the side output $g^{(s)}$ can be formulated as below
\begin{equation}
\frac{\partial \mathcal{L}}{\partial g^{(s)}} = \frac{\partial \mathcal{L}}{\partial g^{(f)}}
\cdot \frac{\partial g^{(f)}}{\partial g^{(s)}}, s=1, 2, \dots, S.
\end{equation}

Second, we formulate the gradient of $\mathcal{L}$ with respect to the inputs of the side branches.
We notate $a^{(s)}$ as the input of the $s$-th side branch.
As depicted in Fig.~\ref{fig:DFN}, $a^{(s)}$ represents the pooling layer.
Note that the input of the main branch, denoted as $a^{(S)}$, refers to the last convolutional layer
(i.e. conv $S$). We can observe that the gradient of $a^{(s)}$ depends on several related branches.
For example in Fig.~\ref{fig:DFN}, the gradient of $a^{(1)}$ is influenced by $S$ of branches;
the gradient of $a^{(2)}$ need to consider the results from the $2$-th to $S$-th branch;
but the gradient of $a^{(S)}$ is updated by only the main branch.
Mathematically, the gradient of $\mathcal{L}$ with respect to the side input $a^{(s)}$ can be computed as follows:
\begin{equation}
\frac{\partial \mathcal{L}}{\partial a^{(s)}} = \sum_{i=s}^{S} \frac{\partial \mathcal{L}}{\partial g^{(i)}}
\cdot \frac{\partial g^{(i)}}{\partial a^{(i)}},
\end{equation}
where $i$ indexes the related branch that contributes to the gradient of $a^{(s)}$.
We then need to sum up the gradients from these related branches.
Like~\cite{Alexnet2012}, we employ standard stochastic gradient descent (SGD) algorithm
with mini-batch to train the whole network.

\subsection{Discussion}
To present more insights into CFN, we compare CFN with other related models.

\textbf{Relationship with CNN.}
Normally, a plain CNN only estimates a final prediction based on the topmost layer, as a result,
the effects of intermediate layers on the prediction are implicit and indirect.
In contrast, CFN can connect the intermediate layers using side branches, and deliver their effects on
the final prediction jointly.
Hence, CFN can take advantage of intermediate layers explicitly and directly.

\textbf{Relationship with DSN.}
DSN~\cite{DSN2015} adds extra supervision to intermediate layers for earlier guidance.
However, CFN that still uses one supervision towards the final prediction aims to generate a fused and richer feature.
In a nutshell, DSN focuses on ``loss fusion'', but CFN instead focuses on ``feature fusion''.

\textbf{Relationship with ResNet.}
ResNet~\cite{ResNet2015} addresses the vanishing gradient problem by adding ``linear'' shortcut connections.
CFN has three main differences as compared to ResNet:
(1) The side branches in CFN are not shortcut connections.
They start from a pooling layer and end in a fusion module together.
(2) In contrast to adding a ``linear'' branch, we still use ReLU in each side branch.
(3) The output of a fusion module is fed to the final prediction.
As mentioned in the ResNet work,
when the network is not overly deep (e.g. 11 or 18 layers),
ResNet may obtain little improvement over a plain CNN.
However, CFN can obtain some considerable improvements as compared to CNN.
In contrast to increasing the depth, CFN can serve as an alternative to improving the discriminative capacity of not-very-deep models.
This explains the usefulness and effectiveness of CFN.

\begin{figure}[t]
\centering
{\includegraphics[width=11cm]{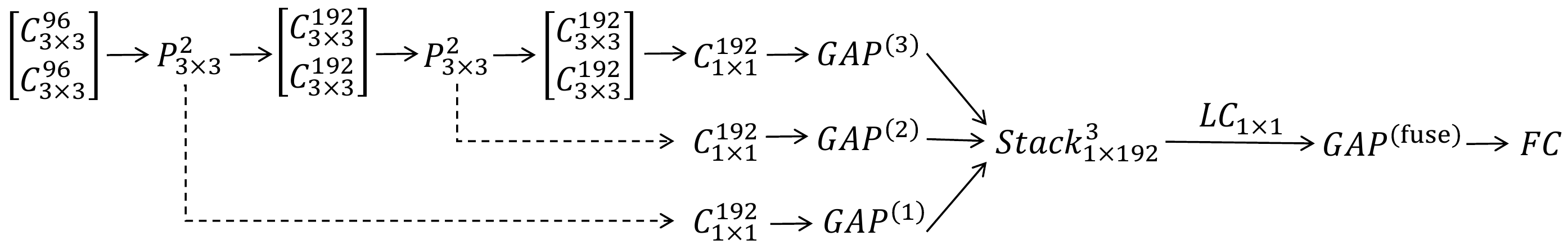} }
\caption{Illustration of CFN built for CIFAR dataset.
For the convolutional layers, the right lower numbers indicate the kernel size;
the right upper number indicates the number of channels.}
\label{fig:modelB}
\end{figure}

\section{Experiments}
First, we trained CFN models on the CIFAR-10/100~\cite{Cifar10} and ImageNet 2012~\cite{Imagenet2015}.
Then, we transferred the trained ImageNet model to three new tasks, including scene recognition, fine-grained recognition and
image retrieval. We conducted all experiments using the Caffe framework~\cite{caffe2014} with a NVIDIA TITAN X card.

\subsection{CIFAR dataset}
Both CIFAR-10~\cite{Cifar10} and CIFAR-100~\cite{Cifar10} consist of 50,000 training images and 10,000 testing images.
But they define 10 and 100 object categories, respectively.
We preprocessed their RGB images by global contrast normalization~\cite{maxout2013}.
We built a plain CNN that consists of seven convolutional layers and one fully-connected layer.
The first six convolutional layers have 3$\times$3 kernel size, but the seventh one is 1$\times$1 convolution.
Global average pooling locates between the last convolutional layer and the fully-connected layer.
Based on the plain CNN, we developed the CFN counterpart as illustrated in Fig.~\ref{fig:modelB}.

Overall, we use the same hyper-parameters to train CNN and CFN.
We use a weight decay of 0.0001, a momentum of 0.9, and a mini-batch size of 100.
The learning rate is initialized with 0.1 and is divided by 10 after $10 \times 10^{4}$ iterations.
The whole training will be terminated after $12 \times 10^{4}$ iterations.
As for CFN, the initialized weights in LC fusion is set with $1/S$ ($S$ is the number of branches).
\begin{table}[b]\small
\captionsetup{belowskip=-10pt}
\centering
\caption{Test error (\%) on CIFAR-10/100 dataset (without data augmentation).}
\begin{tabular}{c|c|cc}
 Model & \#parameters & CIFAR-10 & CIFAR-100 \\ \hline
 CNN & 1.287M (basic)  & 9.28 & 31.89 \\
 CNN-Sum & 1.287M  + 0.074M (extra branches) + 0 (fusion) & 8.84 & 31.42 \\
 CNN-Conv & 1.287M  + 0.074M (extra branches) + 4 (fusion) & 8.68 & 31.16 \\
 CFN & 1.287M + 0.074M (extra branches) + 768 (fusion) & \textbf{8.27} & \textbf{30.68} \\
\end{tabular}
\label{tbl:fusion}
\end{table}

\textbf{Results.}
Table~\ref{tbl:fusion} shows the results on CIFAR-10 test set. We can analyze the results from the following aspects:
(1) Compared with the plain CNN, CFN achieves about 1.01\% and 1.21\% improvement on CIFAR-10 and CIFAR-100, respectively.
(2) In order to demonstrate the advantage of LC fusion, we also implement the
sum-pooling fusion and convolution fusion, denoted as CNN-Sum and CNN-Conv.
We can see that LC fusion used in CFN outperforms both CNN-Sum and CNN-Conv.
(3) We compute the number of parameters in each model. Importantly, the additional parameters for extra side branches
and LC fusion are significantly fewer than the number of basic parameters.
Although LC fusion uses a little more parameters for fusing branches, these parameters are nearly negligible for a deep network.
To reflect the efficiency, we also compare the training time between CNN and CFN.
For example on CIFAR-10, CNN and CFN consumes 1.67 and 2.08 hours, respectively.
\begin{figure}[t]
\setlength{\abovecaptionskip}{0.cm}
\captionsetup{belowskip=-5pt}
\centering
\subfigure[CNN activations] {\includegraphics[height=5.2cm]{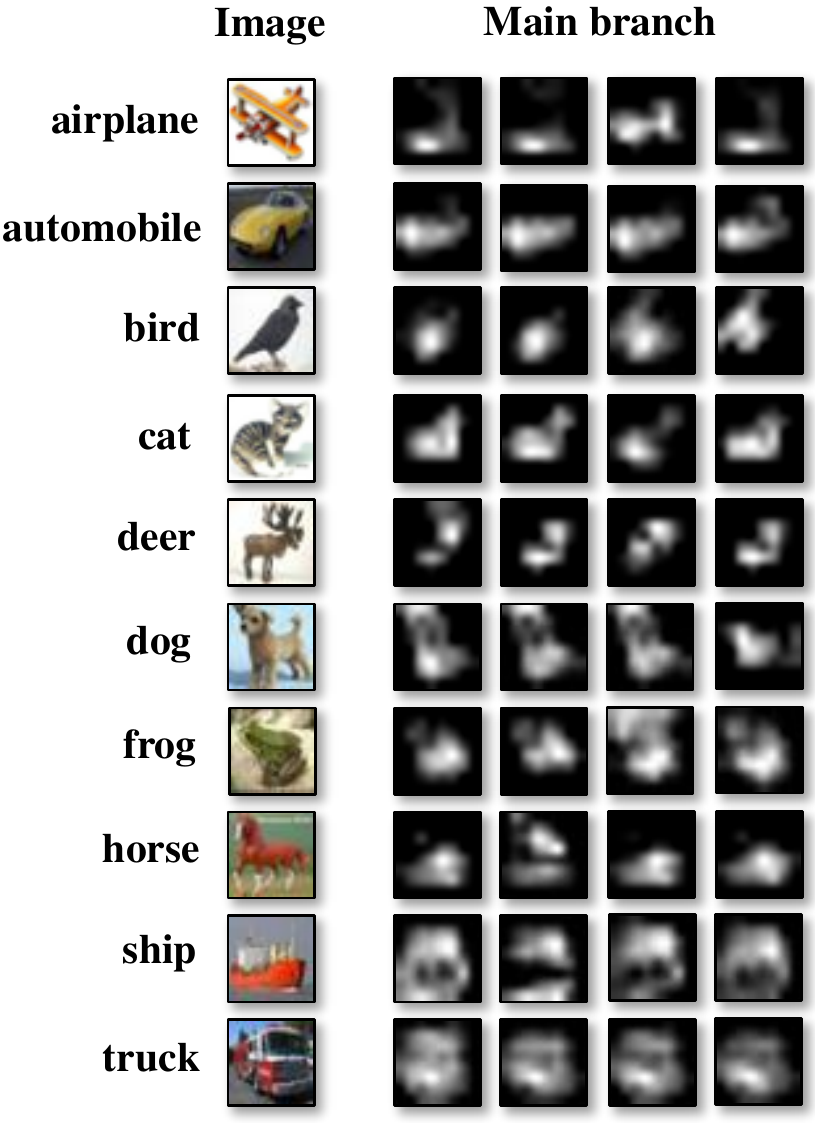} \label{fig:activations_CNN} } \hspace{0.5cm}
\subfigure[CFN activations] {\includegraphics[height=5.2cm]{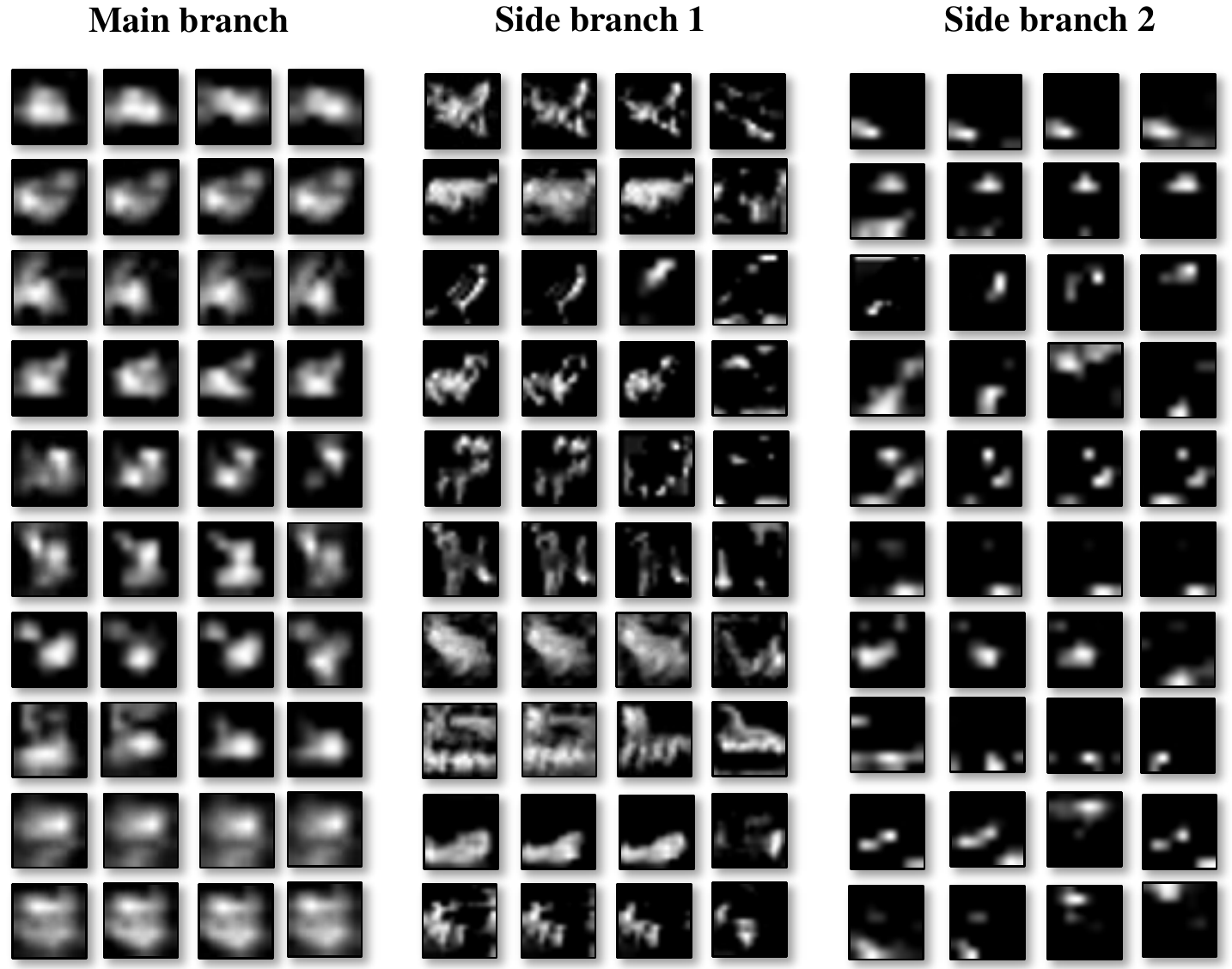} \label{fig:activations_DFN}}
\caption{Illustration of features activations of CIFAR-10 images.
         (a) for CNN, we visualize its 1$\times$1 convolutional layer in the main branch.
         (b) for CFN, the 1$\times$1 convolutional maps from the main branch and two side branches are shown.}
\label{fig:activations}
\end{figure}
\begin{figure}[t]
\setlength{\abovecaptionskip}{0cm}
\captionsetup{belowskip=-10pt}
\centering
\subfigure[CIFAR-10] {\includegraphics[width=5cm,height=3cm]{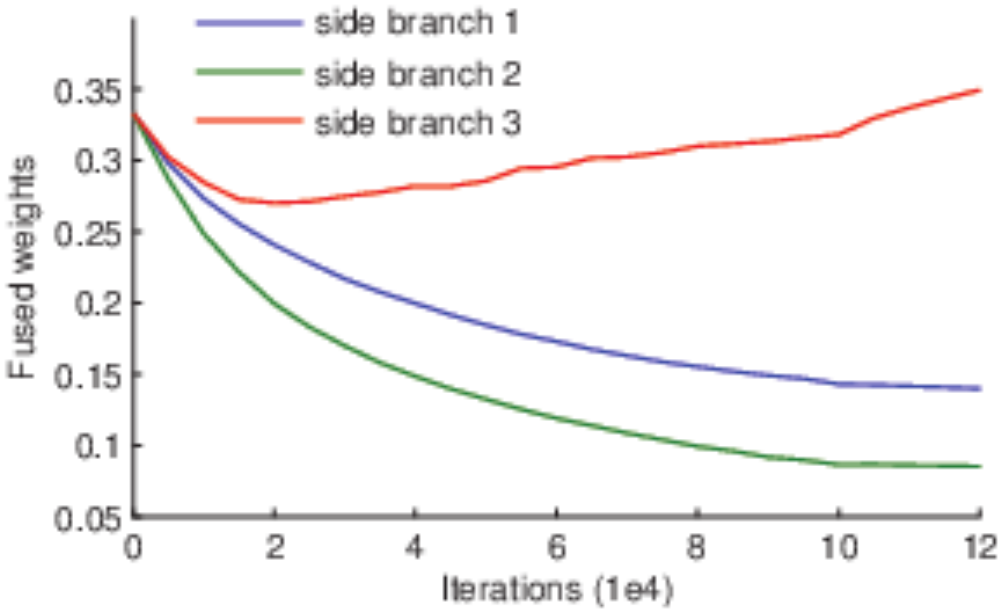} \label{fig:cifar10_weights}}
\subfigure[ImageNet 2012] {\includegraphics[width=6cm,height=3cm]{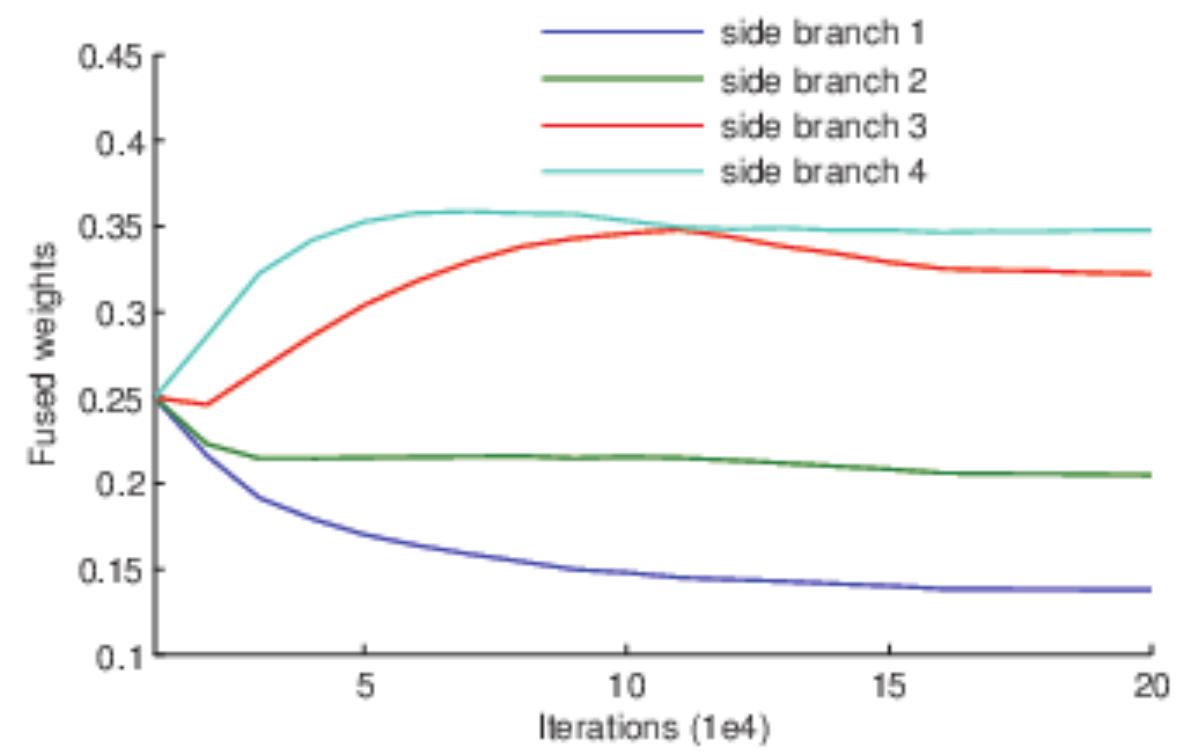} \label{fig:imagenet_weights}}
\caption{Illustration of adaptive weights of the side branches learned in the LC fusion. The top branches have larger weights
than the bottom branches.}
\label{fig:LC_weights}
\end{figure}

In Fig.~\ref{fig:activations}, we visualize and compare the learned feature maps in CNN and CFN.
We select ten images from CIFAR-10 dataset.
We extract the feature maps in the 1$\times$1 convolutional layer and visualize the top-4 maps
(we rank the feature maps by averaging their own activations).
One can observe that CFN can learn complementary clues in the side branches to the full depth main branch.
For example, the side-output 1 mainly learns the boundaries or shapes around the objects.
The side-output 2 focuses on some semantic ``parts'' that fire strong near the objects.
Furthermore, Figure~\ref{fig:cifar10_weights} shows the adaptive weights learned in the LC fusion.
The side branch 3 (main branch) plays a core role,
but other side branches are also complementary to the full depth main branch.

\textbf{Comparison with the state-of-the-art.}
Table~\ref{tbl:comparison} compares the results on CIFAR datasets.
Overall, CFN can obtain comparative results and outperform recent not-very-deep state-of-the-art models.
It is worth mentioning that some work intends to push the results using much deeper networks~\cite{ResNet2015}
and large data augmentation~\cite{Fractional2014}.
In contrast to purely pushing the results, our aim is to demonstrate the advantage of fusing intermediate layers.
Thus we only use a not-overly deep model and standard data augmentation~\cite{DSN2015}.
We believe that Adapting CFN to a very deep model will
be an interesting future work.
\begin{table}[t]
\centering
\caption{Test error on CIFAR-10/100 to compare CFN with recent state-of-the-art.
         A superscripted * indicates to use the standard data augmentation~\cite{DSN2015}.
         }
\begin{tabular}{c|ccc}
 Method & \hspace{0.1cm}CIFAR-10\hspace{0.1cm} & \hspace{0.1cm}CIFAR-10$^{*}$ \hspace{0.1cm}&
  \hspace{0.1cm}CIFAR-100 \hspace{0.1cm}\\ \hline
 Maxout Networks~\cite{maxout2013} & 11.68\% & 9.38\% & 38.57\% \\
 NIN~\cite{NIN2014} & 10.41\% & 8.81\% & 35.68\%  \\
 DSN~\cite{DSN2015} & 9.69\% & 7.97\% & 34.54\%  \\
 ALL-CNN~\cite{AllCNN2015} & 9.08\% & 7.25\% & 33.71\%  \\
 R-CNN~\cite{R-CNN2015} & 8.69\% & 7.09\% & 31.75\%  \\
 NIN + SReLU~\cite{SReLU2016} & 8.41\% & 6.98\% & 31.10\%  \\ \hline
 \textbf{CNN (baseline)} & 9.28\% & 7.34\% & 31.89\%   \\
 \textbf{CFN (ours)} & \textbf{8.27}\% & \textbf{6.77}\% & \textbf{30.68\%}  \\
\end{tabular}
\label{tbl:comparison}
\end{table}
\begin{figure}[b]
\setlength{\abovecaptionskip}{0.cm}
\captionsetup{belowskip=-10pt}
\centering
\includegraphics[width=12cm]{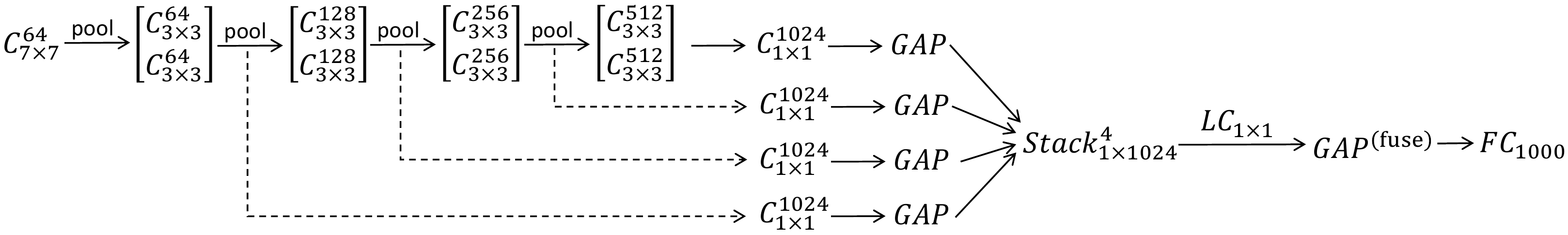}
\caption{The architecture of CFN built for ImageNet classification.}
\label{fig:imagenet}
\end{figure}

\subsection{ImageNet 2012}
We developed a basic 11-layer plain CNN (i.e. CNN-11) whose channels of feature maps range from 64 to 1024.
Based on this CNN, we built its CFN counterpart (i.e. CFN-11) as illustrated in Fig.~\ref{fig:imagenet}.
We create three extra side branches from the pooling layers (excluding the first pooling layer).
Following existing literature~\cite{Alexnet2012,VGGnet2015,googlenet2015,ResNet2015},
we use a weight decay of 0.0001, and a momentum of 0.9 and a mini-batch size of 64, .
Batch normalization (BN)~\cite{BN2015} is used after each convolution.
The learning rate starts from 0.01 and decreases to 0.001 at $10 \times 10^{4}$ iterations,
and to 0.0001 at $15 \times 10^{4}$ iterations.
The whole training will be terminated after $20 \times 10^{4}$ iterations.
LC weights are initialized with 0.25 due to four side branches in total.

\textbf{Results.}
Table~\ref{tbl:imagenet} compares the results on the validation set.
First, CNN-11 can achieve competitive results as compared to AlexNet~\cite{Alexnet2012}, however,
it consumes much fewer parameters ($\sim$6.3 millions) than Alexnet ($\sim$60 millions).
Second, CFN-11 obtains about 1\% improvement over CNN-11, while adding few parameters ($\sim$0.5 millions).
It verifies the efficiency of fusing multi-scale deep representations.
Furthermore, we reproduce the DSN~\cite{DSN2015} and ResNet~\cite{ResNet2015} models based on the plain CNN-11.
As a result, CFN-11 can achieve better accuracy than DSN-11 and ResNet-11.
For such a not-overly deep network, CFN can serve as an alternative to improving the discriminative capacity of CNNs,
instead of increasing the depth like ResNet.
Moreover, to test the generalization of CFN to deeper networks, we build a 19-layer model following the
principle of 11-layer model. Likewise, CFN-19 outperforms
CNN-19 by a consistent improvement as seen in Table~\ref{tbl:imagenet}.

\begin{figure}[t]
\setlength{\abovecaptionskip}{0.cm}
\centering
\includegraphics[width=10.5cm,height=3.5cm]{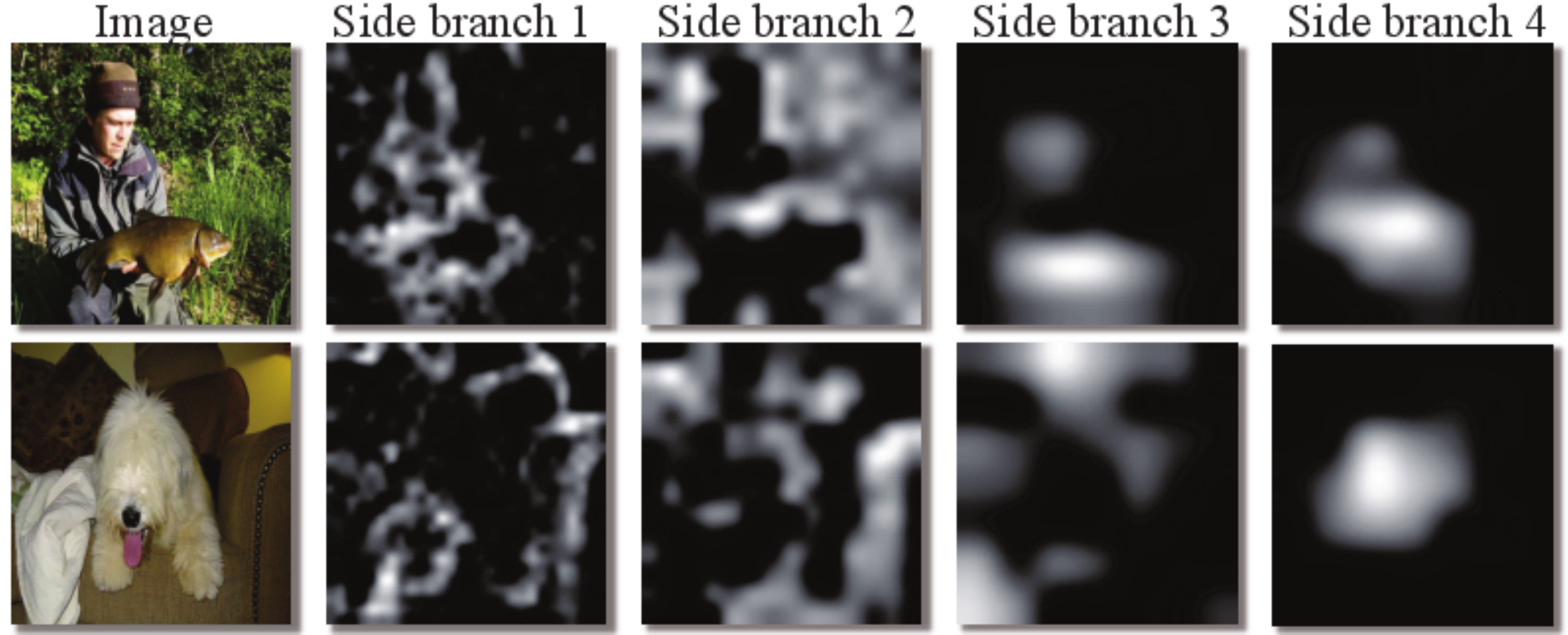}
\caption{Illustration of the feature activations of images of the ImageNet dataset.}
\label{fig:imagenet_activations}
\end{figure}

Similar to CIFAR-10, Figure~\ref{fig:imagenet_weights} shows the adaptive weights learned in the LC fusion.
We can see that the top branches (i.e. 3 and 4) have larger weights than the bottom branches (i.e. 1 and 2).
In Fig.~\ref{fig:imagenet_activations}, we illustrate and compare the feature maps in the side branches.
\begin{table}[t]
\setlength{\abovecaptionskip}{0cm}
\centering
\caption{Error rates (\%) on the ImageNet 2012 validation set.}
\begin{tabular}{c|ccccc|cc} \hline
  \hspace{0.2cm}Method\hspace{0.2cm}& \hspace{0.1cm}AlexNet\hspace{0.1cm} & \hspace{0.1cm}CNN-11\hspace{0.1cm}
  & \hspace{0.1cm}DSN-11\hspace{0.1cm}& \hspace{0.1cm}ResNet-11\hspace{0.1cm}
  & \hspace{0.1cm}CFN-11\hspace{0.1cm} & \hspace{0.1cm}CNN-19\hspace{0.1cm} & \hspace{0.1cm}CFN-19\hspace{0.1cm} \\ \hline
  Top-1& 42.90& 43.11 & 42.24& 43.02& \textbf{41.96}& 36.99 & \textbf{35.47} \\
  Top-5& 19.80& 19.91& 19.24& 19.85& \textbf{19.09} & 14.74 & \textbf{13.93} \\
\end{tabular}
\label{tbl:imagenet}
\end{table}
\begin{table}[b]
\captionsetup{belowskip=-5pt}
\centering
\caption{Results on transferring the ImageNet model to three target tasks.}
\begin{tabular}{c|c|cc|cc|cc}
\multicolumn{2}{c}{} & \multicolumn{2}{c}{Scene recognition} & \multicolumn{2}{c}{Fine-grained recognition} & \multicolumn{2}{c}{Image retrieval} \\ \hline
 \hspace{0.1cm}Method \hspace{0.1cm}& Dim &  \hspace{0.1cm}Scene 15\hspace{0.1cm}&  \hspace{0.1cm}Indoor 67 \hspace{0.1cm}
  &  \hspace{0.1cm}Flower \hspace{0.1cm} &  \hspace{0.1cm}Bird \hspace{0.1cm}  &  \hspace{0.1cm}Holidays \hspace{0.1cm}
   &  \hspace{0.1cm}UKB \hspace{0.1cm}   \\ \hline\hline
 AlexNet~\cite{Alexnet2012}& 4096 &    83.99&	  58.28&   78.68&  45.79&    76.77 & 3.45  \\
 CNN-11& 1024  &  84.32  &	60.45  &   76.79&  45.98&    78.33 & 3.47  \\
 CFN-11& 1024 &  \textbf{86.83}   &    \textbf{62.24}  &   \textbf{82.57}&  \textbf{48.12}&    \textbf{80.32} & \textbf{3.54}  \\
\end{tabular}
\label{tbl:transfer}
\end{table}

\subsection{Transferring fused feature to new tasks}
To evaluate the generalization of CFN, we transferred the trained ImageNet model to three new tasks:
scene recognition, fine-grained recognition and image retrieval.
Each task is evaluated on two datasets: Scene-15~\cite{Scene15} and Indoor-67~\cite{indoor67}, Flower~\cite{Flower2008} and Bird~\cite{Bird2011},
and Holidays~\cite{Holidays2008} and UKB~\cite{UKB2006}.
For AlexNet, the fc7 layer is used as a baseline; For CNN-11, we extract the result of global average pooling as another baseline;
For CFN-11, the fused feature is extracted to represent images.
For scene and fine-grained recognition, we use linear SVM~\cite{libsvm2011} to compute the classification accuracy.
For image retrieval, we use KNN to compute the mAP on Holidays and N-S score on UKB.
Table~\ref{tbl:transfer} reports the evaluation results on six datasets.
We can see that CFN-11 obtains consistent improvement performance on all datasets.
Interestingly, their gains are more remarkable than those in ImageNet.
It reveals that learning multi-scale deep representations are beneficial for diverse vision recognition problems.
In addition, fine-tuning the model on the target datasets will further improve the results.

\section{Conclusions}
We proposed efficient convolutional fusion networks (CFN) by adding few parameters.
It can serve as an alternative to improving recognition accuracy instead of increasing the depth.
Experiments on the CIFAR and ImageNet datasets demonstrate the superiority of CFN over the plain CNN.
Additionally, CFN outperforms not-very-deep state-of-the-art models by considerable gains.
Moreover, we verified its significant generalization while transferring CFN to three new tasks.
In future work, we will evaluate CFN with much deeper neural networks.
\\
\\
\noindent
\textbf{Acknowledgments} 
This work was supported mainly by the LIACS Media Lab at
Leiden University and in part by the China Scholarship Council.
We would like to thank NVIDIA for the donation of GPU cards.

\small
\bibliographystyle{splncs03}
\bibliography{CFN}

\end{document}